%% file: arxiv.tex
\documentclass[10pt,twocolumn, letterpaper]{article}         
\usepackage{cvpr}
\makeatletter
\@namedef{ver@everyshi.sty}{}
\makeatother
\usepackage{times}
\usepackage{epsfig}
\usepackage{graphicx}
\usepackage{amsmath}
\usepackage{amssymb}
\usepackage{tikz}
\usepackage{bm}
\usepackage{pgfplots}
\usepackage{pgf-umlsd}
\usepackage{ifthen}
\usepackage{mathrsfs}
\usepackage{float}
\usepackage{subfigure}
\usepackage{array, boldline, makecell, booktabs}
\usepackage{rotating}
\usetikzlibrary{calc}
\usepackage{multirow}
\newcommand{\doubleconv}[3][]{$\left\{\begin{array}{c}#2\times#2, #3\\#2\times#2, #3\end{array}\right\}#1$}
\newcommand{\resconv}[4]{$\left[\begin{array}{c}1\times1, #1\\ 3\times 3, #2\\1\times 1, #3\end{array}\right] \times #4$}
\newcommand{\deconv}[3]{$\left[\begin{array}{c}1\times1, \pgfmathparse{#1/2}\pgfmathprintnumber{\pgfmathresult}; 1\times1, \pgfmathparse{#1/2}\pgfmathprintnumber{\pgfmathresult}\\ 3\times 3, #2\\1\times 1, #3\end{array}\right]*$}
\newcommand{\insertResults}[4]{
	\begin{minipage}[c]{#3}
		\foreach \x in #1
		{ 
			\includegraphics[width=1\linewidth]{figures/results_low_res/\x_#2}
		}
		\centering #4
	\end{minipage}
}
\newcommand{\bfs}[1]{{\bf #1}}
\makeatletter
\renewcommand{\paragraph}{%
	\@startsection{paragraph}{4}%
	{\z@}{1.0ex \@plus 1ex \@minus .2ex}{-1em}%
	{\normalfont\normalsize\bfseries}%
}
\makeatother

\usepackage[pagebackref=true,breaklinks=true,letterpaper=true,colorlinks,bookmarks=false]{hyperref}

\cvprfinalcopy 

\def\cvprPaperID{145} 

\ifcvprfinal\fi
\begin{document}
	\title{Learning Attraction Field Representation for Robust Line Segment Detection}
	\author{{Nan Xue$^{1}$}, 
		{Song Bai$^{2}$},
		{Fudong Wang$^{1}$},
		{Gui-Song Xia$^{1}$},
		{Tianfu Wu$^{3}$,
			{Liangpei Zhang$^{1}$}}
		\vspace{2mm}\\
		{$^1$State Key Lab. LIESMARS, Wuhan University, China}
		\\{\small \tt \{xuenan, fudong-wang, guisong.xia, zlp62\}@whu.edu.cn}
		\\
		{$^2$Dept. Engineering Science, University of Oxford, UK.} 
		\\{\small \tt songbai.site@gmail.com}	 
		\\
		{$^3$Dept. Electrical \& Computer Engineering, NC State University, USA}
		\\{\small \tt tianfu\_wu@ncsu.edu}
	}
	\maketitle
	\begin{abstract}
		This paper presents a region-partition based attraction field dual representation for line segment maps, and thus poses the problem of line segment detection (LSD) as the region coloring problem. The latter is then addressed by learning deep convolutional neural networks (ConvNets) for accuracy, robustness and efficiency. For a 2D line segment map, our dual representation consists of three components: (i) A region-partition map in which every pixel is assigned to one and only one line segment; (ii) An attraction field map in which every pixel in a partition region is encoded by its 2D projection vector w.r.t. the associated line segment; and (iii) A squeeze module which squashes the attraction field to a line segment map that almost perfectly recovers the input one. By leveraging the duality, we learn ConvNets to compute the attraction field maps for raw input images, followed by the squeeze module for LSD, in an end-to-end manner. Our method rigorously addresses several challenges in LSD such as local ambiguity and class imbalance. Our method also harnesses the best practices developed in ConvNets based semantic segmentation methods such as the encoder-decoder architecture and the a-trous convolution. In experiments, our method is tested on the WireFrame dataset~\cite{Huang2018a} and the YorkUrban dataset~\cite{Denis2008} with state-of-the-art performance obtained. Especially, we advance the performance by $4.5$ percents on the WireFrame dataset. Our method is also fast with $6.6\sim 10.4$ FPS, outperforming most of the existing  line segment detectors. The source code of our method is available at \url{https://github.com/cherubicXN/afm_cvpr2019}.
	\end{abstract}
	
	\vspace{-2mm}
	\section{Introduction}
	\begin{figure}
		\centering
		\subfigure[Attraction field map representation for line segments\label{fig:1a}]{
			\includegraphics[width=.977\linewidth]{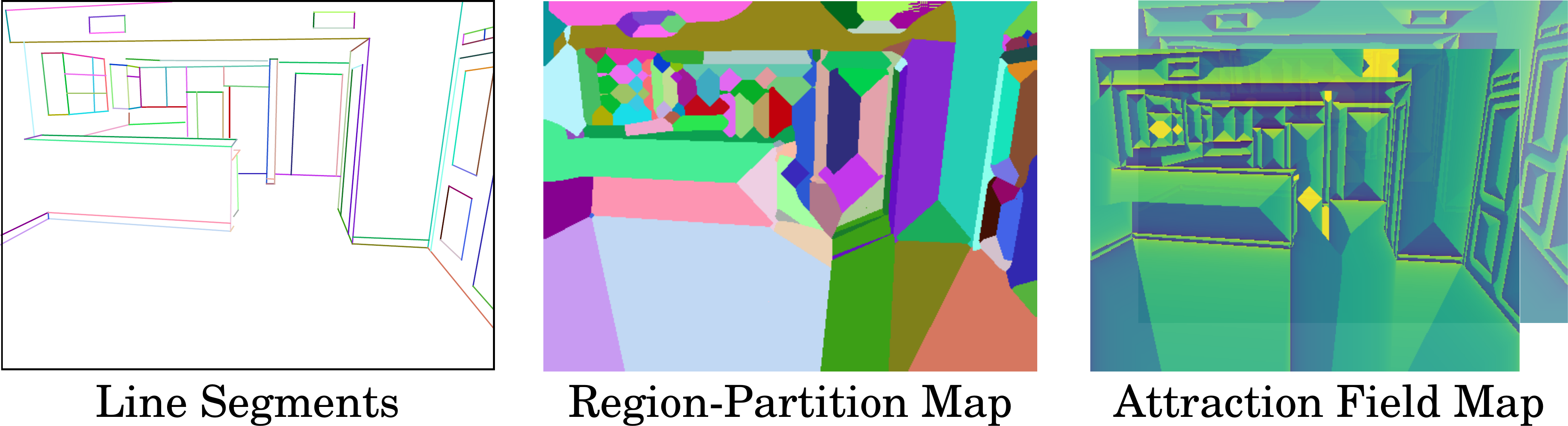}} 
		\subfigure[Our approach for line segment detection\label{fig:1b}]{
			\includegraphics[width=.977\linewidth]{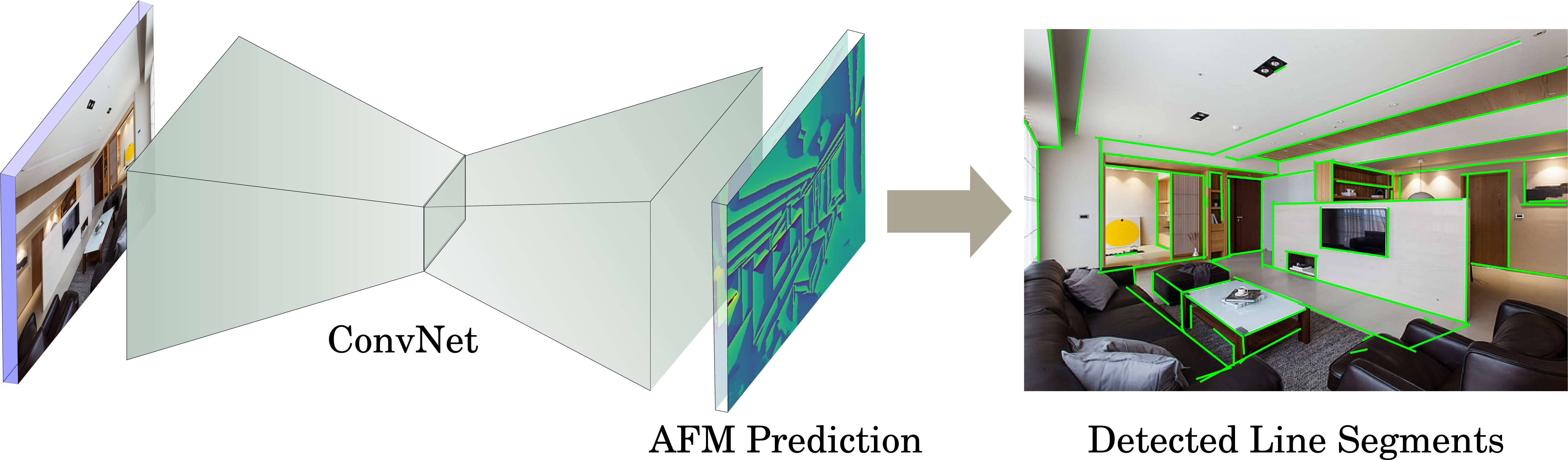}}
		\caption{Illustration of the proposed method. (a) The proposed attraction field dual representation for line segment maps. A line segment map can be almost perfectly recovered from its attraction filed map (AFM), by using a simple squeeze algorithm. (b) The proposed formulation of posing the LSD problem as the region coloring problem. The latter is addressed by learning ConvNets.}
		\label{fig:regional-representation}
	\end{figure}
	
	\subsection{{Motivation and Objective}}
	{Line segment detection (LSD) is an important yet challenging low-level task in computer vision. The resulting line segment maps provide compact structural information that facilitate many up-level vision tasks such as 3D reconstruction~\cite{Denis2008,FaugerasDMAR92}, image partition~\cite{Duan2015}, stereo matching~\cite{Yu2013}, scene parsing~\cite{Zoua,Zhao2013}, camera pose estimation~\cite{Xu2017}, and image stitching~\cite{XiangXBZ18}.} 
	
	{LSD usually consists of two steps: line heat map generation and line segment model fitting. The former can be computed either simply by the gradient magnitude map (mainly used before the recent resurgence of deep learning) \cite{VonGioi2010, Xue2017, Cho2018}, or by a learned convolutional neural network (ConvNet) \cite{Xie2015a, Maninis2016} in state-of-the-art methods \cite{Huang2018a}. The latter needs to address the challenging issue of handling unknown multi-scale discretization nuisance factors ({\em e.g.,} the classic zig-zag artifacts of line segments in digital images) when aligning pixels or linelets to form line segments in the line heat map. Different schema have been proposed, {\em e.g.,} the $\epsilon$-meaningful alignment method proposed in~\cite{VonGioi2010} and the junction~\cite{Xia2014} guided alignment method proposed in~\cite{Huang2018a}. The main drawbacks of existing two-stage methods are in two-fold: lacking elegant solutions to solve the local ambiguity and/or class imbalance in line heat map generation, and requiring extra carefully designed heuristics or supervisedly learned contextual information in inferring line segments in the line heat map.}
	
	In this paper, we focus on learning based LSD framework and propose a single-stage method which rigorously addresses the drawbacks of existing LSD approaches. Our method is motivated by two observations, 
	\begin{itemize}
		\item The duality between region representation and boundary contour representation of objects or surfaces, which is a well-known fact in computer vision.
		\item The recent remarkable progresses for image semantic segmentation by deep ConvNet based methods such as U-Net \cite{Ronneberger2015} and DeepLab V3+ \cite{Hou2018}.  
	\end{itemize} 
	So, the intuitive idea of this paper is that if we can bridge line segment maps and their dual region representations, we will pose the problem of LSD as the problem of region coloring, and thus open the door to leveraging the best practices developed in  state-of-the-art deep ConvNet based image semantic segmentation methods to improve performance for LSD. By dual region representations, it means they are capable of recovering the input line segment maps in a nearly perfect way via a simple algorithm. We present an efficient and straightforward method for computing the dual region representation. By re-formulating LSD as the equivalent region coloring problem, we address the aforementioned challenges of handling local ambiguity and class imbalance in a principled way.
	
	\subsection{Method Overview}
	Figure~\ref{fig:regional-representation} illustrates the proposed method. Given a 2D line segment map,
	we represent each line segment by its geometry model using the two end-points\footnote{We will have discrepancy for some intermediate points of a line segment between their annotated pixel locations and the geometric locations when the line segment is not strictly horizontal or vertical.}. In computing the dual region representation, there are three components (detailed in Section~\ref{sec:representation}). 
	\begin{itemize}
		\item A region-partition map. It is computed by assigning every pixel to one and only one line segment based on a proposed point to line segmentation distance function. The pixels associated with one line segment form a region. All regions represent a partition of the image lattice ({\em i.e.,} mutually exclusive and the union occupies the entire image lattice).
		\item An attraction field map. Each pixel in a partition region has one and only one corresponding projection point on the geometry line segment (but the reverse is often  a one-to-many mapping). In the attraction field map, every pixel in a partition region is then represented by its attraction/projection vector between the pixel and its projection point on the geometry line segment \footnote{They are the same point when the pixel is on the geometry line segment, and thus we will have a zero vector. We observed that the total number of those points are negligible in our experiments.}.  
		\item A light-weight squeeze module. It follows the attraction field to squash partition regions in an attraction field map to line segments that almost perfectly recovers the input ones, thus bridging the duality between region-partition based attraction field maps and line segment maps. 
	\end{itemize} 
	
	The proposed method can also be viewed as an intuitive expansion-and-contraction operation between 1D line segments and 2D regions in a simple projection vector field: The region-partition map generation jointly expands all line segments into partition regions, and the squeeze module degenerates regions into line segments. 
	
	With the duality between a line segment map and the corresponding region-partition based attraction field map, we first convert all line segment maps in the training dataset to their attraction field maps. Then, we learn ConvNets to predict the attraction field maps from raw input images in an end-to-end way. We utilize U-Net \cite{Ronneberger2015} and a modified network based on DeepLab V3+ \cite{Hou2018} in our experiments. After the attraction field map is computed, we use the squeeze module to compute its line segment map.
	
	In experiments, the proposed method is tested on the WireFrame dataset~\cite{Huang2018a} and the YorkUrban dataset~\cite{Denis2008} with state-of-the-art performance obtained comparing with \cite{Huang2018a, Cho2018,Almazan_2017_CVPR, VonGioi2010}. In particular, we improve the performance by $4.5\%$ on the WireFrame dataset. Our method is also fast with  $6.6\sim 10.4$ FPS, outperforming most of line segment detectors.
	
	\section{Related Work and Our Contributions}
	The study of line segment detection has a very long history since 1980s \cite{Ballard81}.
	The early pioneers tried to detect line segments based upon the edge map estimation. Then, the perception grouping approaches based on the \emph{Gestalt Theory} are proposed. Both of these methods concentrate on the hand-crafted low-level features for the detection, which have become a limitation. Recently, the line segment detection and its related problem edge detection have been studied under the perspective of deep learning, which dramatically improved the detection performance and brings us of great practical importance for real applications. 
	
	\subsection{Detection based on Hand-crafted Features}
	In a long range of time, the hand-crafted low-level  features (especially for image gradients) are heavily used for line segment detection. These approaches can be  divided into edge map based approaches \cite{FurukawaS03,Kamat-SadekarG98,XuSK15,XuSK15A,XuSK15B, Almazan_2017_CVPR} and perception grouping approaches \cite{BurnsHR86,VonGioi2010, Cho2018}. The edge map based approaches treat the visual features as a discriminated feature for edge map estimation and subsequently applying the Hough transform \cite{Ballard81} to globally search line configurations and then cutting them by using thresholds. In contrast to the edge map based approaches, the grouping methods directly use the image gradients as local geometry cues to group pixels into line segment candidates and filter out the false positives \cite{VonGioi2010, Cho2018}.
	
	Actually, the features used for line segment detection can only characterize the local response from the image appearance. For the edge detection, only local response without global context cannot avoid false detection. On the other hand, both the magnitude and orientation of image gradients are easily affected by the external imaging condition ({\em e.g.} noise and illumination). Therefore, the local nature of these features limits us to extract line segments from images robustly. In this paper, we break the limitation of locally estimated features and turn to learn the deep features that hierarchically represent the information of images from low-level cues to high-level semantics.
	
	\subsection{Deep Edge and Line Segment Detection}
	Recently,  HED \cite{Xie2015a} opens up a new era for edge perception from images by using ConvNets. The learned multi-scale and multi-level features dramatically addressed the problem of false detection in the edge-like texture regions and approaching human-level performance on the BSDS500 dataset \cite{Martin2004a}. Followed by this breakthrough, a tremendous number of deep learning based edge detection approaches are proposed \cite{Maninis2016,Kokkinos2016,Liu2017a,Kokkinos2017a,Maninis2018a,Hou2018}. 
	Under the perspective of binary classification, the edge detection has been solved to some extent. 
	It is natural to upgrade the traditional edge map based line segment detection by alternatively using the edge map estimated by ConvNets. 
	However, the edge maps estimated by ConvNets are usually over-smoothed, which will lead to local ambiguities for accurate localization. Further, the edge maps do not contain enough geometric information for the detection. 
	According to the development of deep learning, it should be more reasonable to propose an end-to-end line segment detector instead of only applying the advances of deep edge detection.
	
	Most recently, Huang {\em et al.} \cite{Huang2018a} have taken an important step towards this goal by proposing a large-scale dataset with high quality \emph{line segment annotations} and approaching the problem of line segment detection as two parallel tasks, \ie, {edge map detection} and {junction detection}. 
	As a final step for the detection, the resulted edge map and junctions are fused to produce line segments. To the best of our knowledge, this is the first attempt to develop a deep learning based line segment detector. However, due to the sophisticated relation between edge map and junctions, it still remains a problem unsolved. Benefiting from our proposed formulation, we can directly learn the line segments from the attraction field maps that can be easily obtained from the line segment annotations without the junction cues.
	
	\paragraph{Our Contributions} The proposed method makes the following main contributions to robust line segment detection. 
	\begin{itemize}
		\item A novel dual representation is proposed by bridging line segment maps and region-partition-based attraction field maps. To our knowledge, it is the first work that utilizes this simple yet effective representation in LSD.
		\item With the proposed dual representation, the LSD problem is re-formulated as the region coloring problem, thus opening the door to leveraging state-of-the-art semantic segmentation methods in addressing the challenges of local ambiguity and class imbalance in existing LSD approaches in a principled way.
		\item The proposed method obtains state-of-the-art performance on two widely used LSD benchmarks, the WireFrame dataset (with $4.5\%$ significant improvement) and the YorkUrban dataset.  
	\end{itemize}
	
	\section{The Attraction Field Representation}\label{sec:representation}
	In this section, we present details of the proposed region-partition representation for LSD.
	
	\begin{figure}[!b]
		\centering
		\subfigure[Support regions\label{fig:pols-proposal}]{
			\input{figures/support_region}
		}
		\subfigure[Attraction vectors \label{fig:details}]{
			\input{figures/offset_details}
		}
		\subfigure[Squeeze module\label{fig:squeeze}]{
			\input{figures/squeeze}
		}
		\caption{A toy example illustrating a line segment map with 3 line segments, its dual region-partition map, selected vectors of the attraction field map and the squeeze module for obtaining line segments from the attraction field map. See text for details.}
		\label{fig:lsmap}
	\end{figure}
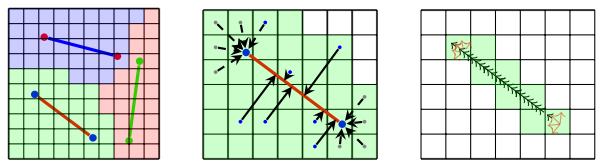
	
	\subsection{The Region-Partition Map}
	Let $\Lambda$ be an image lattice ({\em e.g.}, $800\times 600$). A line segment is denote by $l_i=(\mathbf{x}^s_i, \mathbf{x}^e_i)$ with the two end-points being $\mathbf{x}^s_i$ and $\mathbf{x}^e_i$  (non-negative real-valued positions due to sub-pixel precision is used in annotating line segments) respectively. The set of line segments in a 2D line segment map is denoted by $L=\{l_1, \cdots, l_n\}$ . For simplicity, we also denote the line segment map by $L$.  Figure~\ref{fig:lsmap} illustrates a line segment map with 3 line segments in a $10\times 10$ image lattice. 
	
	Computing the region-partition map for $L$ is assigning every pixel in the lattice to one and only one of the $n$ line segments. To that end, we utilize the point-to-line-segment distance function. Consider a pixel $p\in \Lambda$ and a line segment $l_i=(\mathbf{x}_i^s, \mathbf{x}_i^e)\in L$, we first project the pixel $p$ to the straight line going through $l_i$ in the continuous geometry space. If the projection point is not on the line segment, we use the closest end-point of the line segment as the projection point. Then, we compute the Euclidean distance between the pixel and the projection point. Formally, we define the distance between $p$ and $l_i$ by
	\begin{equation}
	\begin{split}
	d(p, l_i) &= \min_{t\in [0,1]}||\mathbf{x}_i^s + t\cdot (\mathbf{x}_i^e - \mathbf{x}_i^s) - p||_2^2, \\
	t_p^* &= \arg\min_t d(p, l_i),
	\end{split}
	\end{equation}
	where the projection point is the original point-to-line projection point if $t^*_p\in(0, 1)$, and the closest end-point if $t^*_p=0$ or $1$.  
	
	So, the region in the image lattice for a line segment $l_i$ is defined by
	\begin{equation}
	R_i =\{p\, |\, p\in \Lambda; d(p, l_i) < d(p, l_j), \forall j\neq i, l_j\in L\}.
	\end{equation}
	It is straightforward to see that $R_i\cap R_j=\emptyset$ and $\cup_{i=1}^n R_i=\Lambda$, {\em i.e.}, all $R_i$'s form a partition of the image lattice. Figure~\ref{fig:pols-proposal} illustrates the partition region generation for a line segment in the toy example (Figure~\ref{fig:lsmap}). Denote by $R=\{R_1, \cdots, R_n\}$ the \emph{region-partition map} for a line segment map $L$. 
	
	\subsection{Computing the Attraction Field Map}
	Consider the partition region $R_i$ associated with a line segment $l_i$, for each pixel $p\in R_i$, its projection point $p'$ on $l_i$ is defined by
	\begin{equation}
	p' = \mathbf{x}_i^s + t_p^*\cdot (\mathbf{x}_i^e - \mathbf{x}_i^s), 
	\end{equation}
	
	We define the 2D attraction or projection vector for a pixel $p$ as, 
	\begin{equation}
	\mathbf{a}(p) = p' - p,
	\end{equation}
	where the attraction vector is perpendicular to the line segment if $t^*_p\in(0, 1)$ (see Figure~\ref{fig:details}). Figure~\ref{fig:regional-representation} shows examples of the $x$- and $y$-component of an \emph{attraction field map} (AFM). Denote by $A=\{\mathbf{a}(p)\, |\, p\in \Lambda\}$ the attraction field map for a line segment map $L$.
	
	\subsection{The Squeeze Module} 
	Given an attraction field map $A$, we first reverse it by computing the real-valued projection point for each pixel $p$ in the lattice,
	\begin{equation}
	v(p) = p+\mathbf{a}(p), 
	\end{equation}
	and its corresponding discretized point in the image lattice,
	\begin{equation}
	v_{\Lambda}(p) = \lfloor v(p) + 0.5 \rfloor. 
	\end{equation}
	where $\lfloor \cdot \rfloor$ represents the floor operation, and $v_{\Lambda}(p)\in \Lambda$. 
	
	Then, we compute a line proposal map in which each pixel $q\in \Lambda$ collects the attraction field vectors whose discretized projection points are $q$. The candidate set of attraction field vectors collected by a pixel $q$ is then defined by 
	\begin{equation}
	\mathcal{C}(q) = \{\bfs{a}(p) \, | \, p\in \Lambda, v_{\Lambda}(p)=q \}, 
	\end{equation}
	where $\mathcal{C}(q)$'s are usually non-empty for a sparse set of pixels $q$'s which correspond to points on the line segments. An example of the line proposal map is shown in Figure~\ref{fig:squeeze}, which project the pixels of the support region for a line segment into pixels near the line segment.
	
	With the line proposal map, our squeeze module utilizes an iterative and greedy grouping algorithm to fit line segments, similar in spirit to the region growing algorithm used in~\cite{VonGioi2010}. \begin{itemize}
		\item Given the current set of active pixels each of which has a non-empty candidate set of attraction field vectors, we randomly select a pixel $q$ and one of its attraction field vector $\bfs{a}(p)\in \mathcal{C}(q)$. The tangent direction of the selected attraction field vector $\bfs{a}(p)$ is used as the initial direction of the line segment passing the pixel $q$. 
		\item Then, we search the local observation window centered at $q$ ({\em e.g.,} a $3\times 3$ window is used in this paper) to find the attraction field vectors which are aligned with $\bfs{a}(p)$ with angular distance less than a threshold $\tau$ ({\em e.g.,} $\tau=10^\circ$ used in this paper). 
		\begin{itemize}
			\item If the search fails, we discard $\bfs{a}(p)$ from $\mathcal{C}(q)$, and further discard the pixel $q$ if $\mathcal{C}(q)$ becomes empty.
			\item Otherwise, we grow $q$ into a set and update its direction by averaging the aligned attraction  vectors. The aligned attractiion  vectors will be marked as used (and thus inactive for the next round search). For the two end-points of the set, we recursively apply the greedy search algorithm to grow the line segment.
		\end{itemize}
		\item Once terminated, we obtain a candidate line segment $l_q=(\mathbf{x}_q^s, \mathbf{x}_q^e)$ with the support set of real-valued projection points. We fit the minimum outer rectangle using the support set. We verify the candidate line segment by checking the aspect ratio between width and length of the approximated rectangle with respect to a predefined threshold to ensure the approximated rectangle is ``thin enough". If the checking fails, we mark the pixel $q$ inactive and release the support set to be active again.   
	\end{itemize}
	
	\begin{figure}[!h]
		\centering
		\includegraphics[width=.4\linewidth]{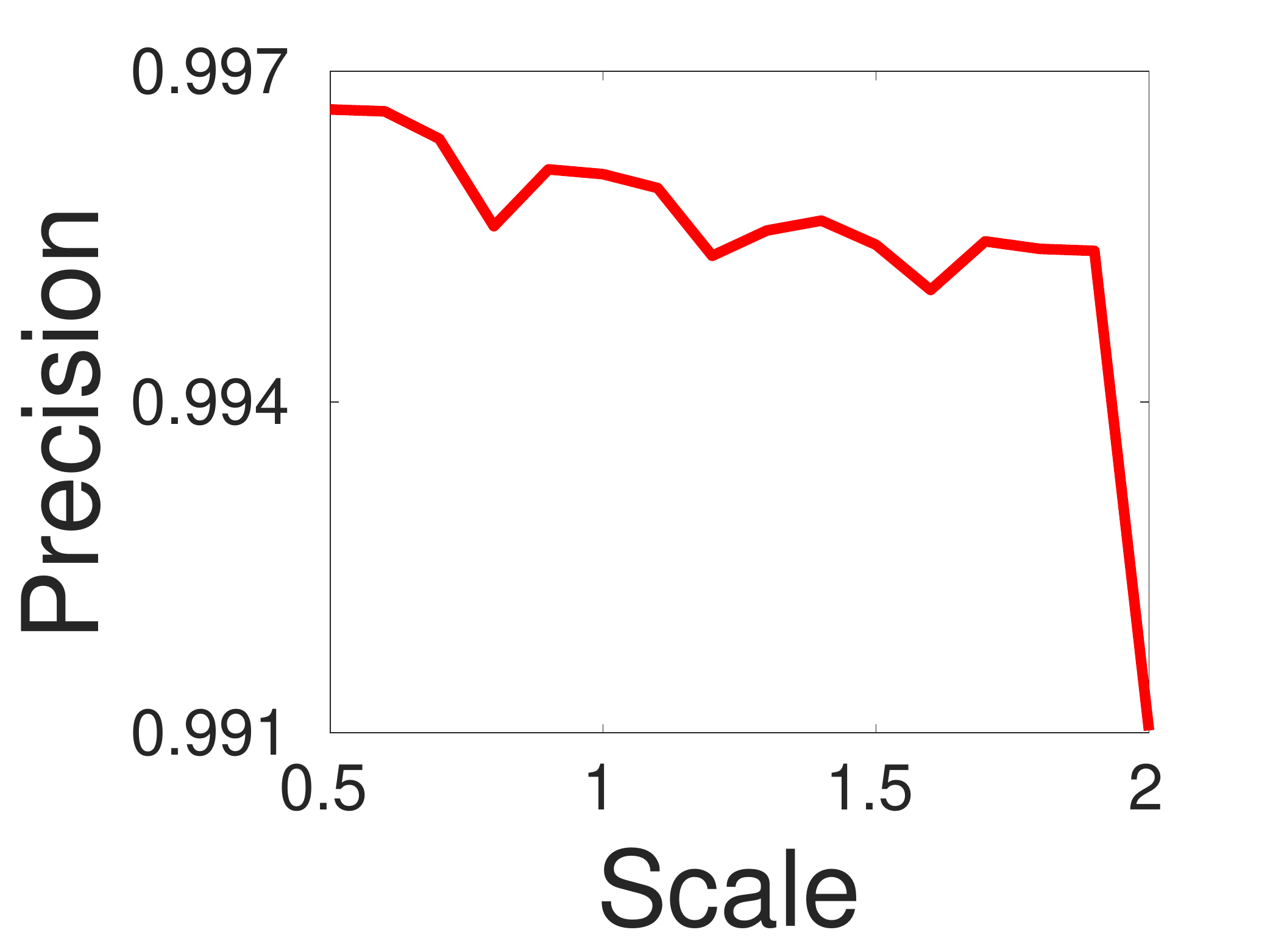}
		\includegraphics[width=.4\linewidth]{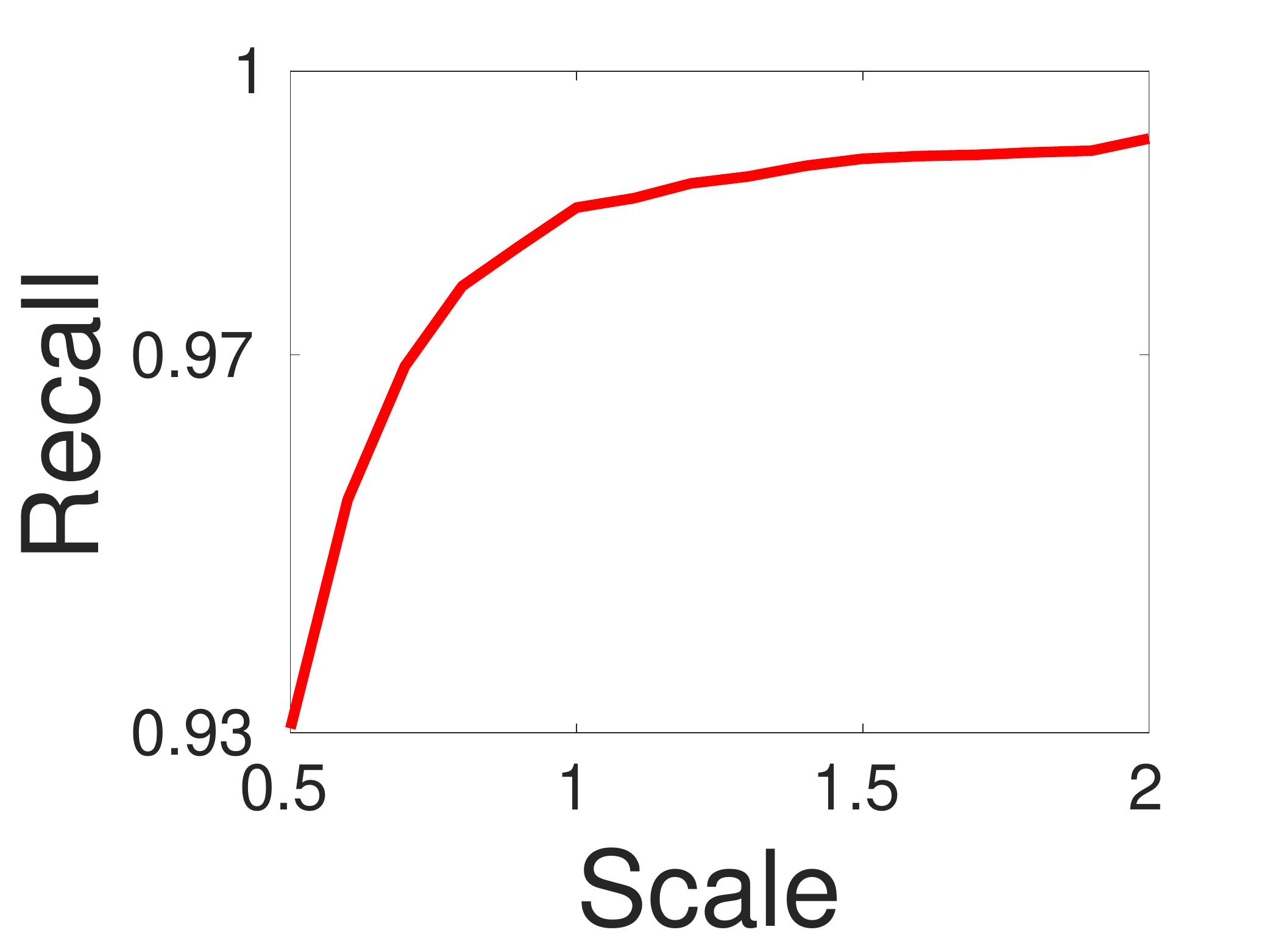}
		\caption{Verification of the duality between line segment maps and attraction field maps, and its scale invariance.}
		\vspace{-4mm}
		\label{fig:scaled-pr}
	\end{figure}
	
	\subsection{Verifying the Duality and its Scale Invariance} 
	We test the proposed attraction field representation on the WireFrame dataset~\cite{Huang2018a}. We first compute the attraction field map for each annotated line segment map and then compute the estimated line segment map using the squeeze module. We run the test across multiple scales, ranging from $0.5$ to $2.0$ with step-size $0.1$.  We evaluate the estimated line segment maps by measuring the precision and recall following the protocol provided in the dataset. Figure~\ref{fig:scaled-pr} shows the precision-recall curves. The average precision and recall rates are above $0.99$ and $0.93$ respectively, thus verifying the duality between line segment maps and corresponding region-partition based attractive field maps, as well as the scale invariance of the duality.
	
	So, \textbf{the problem of LSD can be posed as the region coloring problem almost without hurting the performance.} In the region coloring formulation, our goal is to learn ConvNets to infer the attraction field maps for input images. The attraction field representation  eliminates local ambiguity in traditional gradient magnitude based line heat map, and the predicting attraction field in learning gets rid of the imbalance problem in line \emph{v.s.} non-line classification.
	
	\section{Robust Line Segment Detector}\label{sec:detection}
	In this section, we present details of learning ConvNets for robust LSD. ConvNets are used to predict AFMs from raw input images under the image-to-image transformation framework, and thus we adopt encoder-decoder network architectures. 
	
	\subsection{Data Processing} 
	Denote by $D=\{ (I_i, L_i); i=1,\cdots, N\}$ the provided training dataset consisting of $N$ pairs of raw images and annotated line segment maps. We first compute the AFMs for each training image. Then, let $D=\{ (I_i, \mathbf{a}_i); i=1,\cdots, N\}$ be the dual training dataset. To make the AFMs insensitive to the sizes of raw images, we adopt a simple normalization scheme. For an AFM  $\bfs{a}$ with the spatial dimensions being $W\times H$, the size-normalization is done by
	\begin{equation}
	\bfs{a}_x := \bfs{a}_x/W,~~ \bfs{a}_y := \bfs{a}_y/H,
	\end{equation}
	where $\bfs{a}_x$ and $\bfs{a}_y$ are the component of $\bfs{a}$ along $x$ and $y$ axes respectively. However, the size-normalization will make the values in $\bfs{a}$ small and thus numerically unstable in training. We apply a point-wise invertible value stretching transformation for the size-normalized AFM 
	\begin{equation}
	z' := S(z) = -{\rm sign}(z)\cdot\log(|z|+\varepsilon),
	\end{equation}
	where $\varepsilon = 1\mathrm{e}{-6}$ to avoid $\log(0)$. The inverse function $S^{-1}(\cdot)$ is defined by
	\begin{equation}
	\label{eq:inverse-mapping}
	z := S^{-1}(z') = {\rm sign}(z') e^{(-|z'|)}.
	\end{equation} 
	
	For notation simplicity, denote by $R(\cdot)$ the composite reverse function, and we still denote by $D=\{ (I_i, \mathbf{a}_i); i=1,\cdots, N\}$ the final training dataset. 
	\subsection{Inference} Denote by $f_{\Theta}(\cdot)$ a ConvNet with the parameters collected by $\Theta$. As illustrated in Figure~\ref{fig:1b}, for an input image $I_{\Lambda}$, our robust LSD is defined by 
	\begin{align}
	\hat{\mathbf{a}} & = f_{\Theta}(I_{\Lambda}) \\
	\hat{L} & = Squeeze(R(\hat{\mathbf{a}}))
	\end{align}
	where $\hat{\mathbf{a}}$ is the predicted AFM for the input image (the size-normalized and value-stretched one), $Squeeze(\cdot)$ the squeeze module and $\hat{L}$ the inferred line segment map. 
	\subsection{Network Architectures} 
	We utilize two network architectures to realize $f_{\Theta}()$: one is  U-Net~\cite{Ronneberger2015}, and the other is a modified U-Net, called \emph{a-trous Residual U-Net} which uses the ASSP module proposed in DeepLab v3+ \cite{Chena} and the skip-connection as done in ResNet \cite{He2016}. 
	
	\begin{table}[t!]
		\centering
		\small
		\caption{Network architectures we investigated for the \emph{attraction field learning}. $\{\}$ and $[]$ represent the double conv in U-Net and the residual block. Inside the brackets are the shape of convolution kernels. The suffix $*$ represent the bilinear upsampling operator with the scaling factor $2$. The number outside the brackets is the number of stacked blocks on a stage.}
		\label{tab:net-arch}
		\begin{tabular}{c|c|c}
			\hline
			stage  & U-Net & \emph{a-trous} Residual U-Net\\\hline
			c1 &  \doubleconv{3}{64} & $3\times3, 64$, stride 1  \\\hline
			\multirow{4}{*}{c2} &   $2\times 2$ max pool, stride 2 & $3\times 3$ max pool, stride 2 \\\cline{2-3}
			&  \doubleconv{3}{128} & \resconv{64}{64}{256}{3} \\\hline
			\multirow{3}{*}{c3} &   $2\times 2$ max pool, stride 2 &  \multirow{2}{*}{\resconv{128}{128}{512}{4}} \\\cline{2-2}
			&  \doubleconv{3}{256} &  \\\hline
			\multirow{3}{*}{c4} &   $2\times 2$ max pool, stride 2 & \multirow{2}{*}{\resconv{256}{256}{1024}{6}} \\\cline{2-2}
			&  \doubleconv{3}{512} &  \\\hline
			\multirow{3}{*}{c5} &   $2\times 2$ max pool, stride 2 & \multirow{2}{*}{ \resconv{512}{512}{2048}{3}} \\\cline{2-2}
			&  \doubleconv{3}{512} & \\\hline
			\multirow{4}{*}{d4} &  \multirow{4}{*}{\doubleconv[*]{3}{256}} & ASPP\\\cline{3-3}
			&   & \deconv{512}{512}{512}\\\hline
			d3 &  \doubleconv[*]{3}{128} & \deconv{256}{256}{256}\\\hline
			d2 & \doubleconv[*]{3}{64} & \deconv{128}{128}{128}\\\hline
			d1 & \doubleconv[*]{3}{64} & \deconv{64}{64}{64} \\\hline
			output & \multicolumn{2}{c}{$1\times 1$, stride 1, w.o. BN and ReLU}
			\\\hline
		\end{tabular}
	\end{table}
	Table~\ref{tab:net-arch} shows the configurations of the two architectures. The network consists of $5$ encoder  and $4$ decoder stages indexed by $c1, \ldots, c5$ and $d1, \ldots, d4$ respectively.
	\begin{itemize}
		\item For U-Net, the \emph{double conv} operator that contains two convolution layers is applied and denoted as $\{\cdot\}$. The $\{\cdot\}*$ operator of $d_i$ stage upscales the output feature map of its last stage and then concatenate it with the feature map of $c_i$ stage together before applying the \emph{double conv} operator. 
		\item For the \emph{a-trous Residual U-Net}, we replace the \emph{double conv} operator to the \emph{Residual block}, denoted as $[\cdot]$. Different from the ResNet, we use the plain convolution layer with $3\times 3$ kernel size and stride $1$. Similar to $\{\cdot\}*$, the operator $[\cdot]*$ also takes the input from two sources and upscales the feature of first input source. The first layer of $[\cdot]*$ contains two parallel convolution operators to reduce the depth of feature maps and then concatenate them together for the subsequent calculations. In the stage $d_4$, we apply the 4 ASPP operators with the output channel size $256$ and the dilation rate $1,6,12,18$ and then concatenate their outputs. The output stage use the  convolution operator with $1\times 1$ kernel size and stride $1$ without batch normalization \cite{BN-IoffeS15} and ReLU \cite{ReLU-NairH10} for the \emph{attraction field map} prediction.
	\end{itemize}
	
	\subsection{Training} 
	We follow standard deep learning protocol to estimate the parameters $\Theta$. 
	
	\textit{Loss function.} We adopt the $l_1$ loss function in training.
	\begin{equation}
	\ell(\hat{\bfs{a}}, \bfs{a}) = \sum_{(x,y)\in \Lambda} \|{\bf{a}(x,y)}-{\bf{\hat{a}}(x,y)}\|_1.
	\end{equation}
	
	\textit{Implementation details.} We train the two networks (U-Net  and  \emph{a-trous} Residual U-Net) from scratch on the training set of Wireframe dataset \cite{Huang2018a}. Similar to \cite{Huang2018a}, we follow the standard data augmentation strategy to enrich the training samples with image domain operations including mirroring and flipping upside-down. 
	The stochastic gradient descent (SGD) optimizer with momentum $0.9$ and initial learning rates $0.01$ is applied for network optimization. We train these networks with $200$ epochs and the learning rate is decayed with the factor of $0.1$ after every $50$ epochs. In training phase, we resize the images  to $320\times 320$ and then generate the offset maps from resized line segment annotations to form the mini batches. As discussed in Section \ref{sec:representation}, the rescaling step with reasonable factor will not affect the results. The mini-batch sizes for the two networks are $16$ and $4$ respectively due to the GPU memory.
	
	In testing, a test image is also resized to $320\times 320$ as input to the network. Then, we use the squeeze module to convert the \emph{attraction field map} to line segments. Since the line segments are insensitive to scales, we can directly resize them to original image size without loss of accuracy. The squeeze module is implemented with C++ on CPU.
	
	\section{Experiments}
	\label{sec:experiments}
	In this section, we evaluate the proposed line segment detector and make the comparison with existing state-of-the-art line segment detectors \cite{Huang2018a, Cho2018, Almazan_2017_CVPR, VonGioi2010}. As shown below, our proposed line segment detector outperforms these existing methods on the WireFrame dataset \cite{Huang2018a} and YorkUrban dataset \cite{Denis2008}.
	
	\subsection{Datasets and Evaluation Metrics}
	We follow the evaluation protocol from the deep wireframe parser \cite{Huang2018a} to make a comparison. Since we train on the Wreframe dataset \cite{Huang2018a}, it is necessary to evaluate our proposed method on its testing dataset, which includes $462$ images for man-made environments (especially for indoor scenes). To validate the generalization ability, we also evaluate our proposed approach on the YorkUrban Line Segment Dataset \cite{Denis2008}. Afterthat, we also compared our proposed line segment detector on the images fetched from Internet.
	
	All methods are evaluated quantitatively by the \emph{precision} and \emph{recall} as described in \cite{Huang2018a, Martin2004a}. The \emph{precision} rate indicates the proportion of positive detection among all of the detected line segments whereas \emph{recall} reflects the fraction of detected line segments among all in the scene. The detected and ground-truth line segments are digitized to image domain and we define the ``positive detection'' pixel-wised. The line segment pixels within $0.01$ of the image diagonal is regarded as positive. After getting the \emph{precision} (P) and \emph{recall} (R), we compare the performance of algorithms with F-measure
	$F = 2\cdot \frac{P\cdot R}{P+R}$.
	
	\subsection{Comparisons for Line Segment Detection}
	We compare our proposed method with Deep Wireframe Parser\footnote{\label{foot:wireframe}\url{https://github.com/huangkuns/wireframe}}   \cite{Huang2018a},
	Linelet\footnote{\label{foot:linelet}\url{https://github.com/NamgyuCho/Linelet-code-and-YorkUrban-LineSegment-DB}} \cite{Cho2018}, 
	the Markov Chain Marginal Line Segment Detector\footnote{\label{foot:mcmlsd}\url{http://www.elderlab.yorku.ca/resources/}}(MCMLSD) \cite{Almazan_2017_CVPR} and the Line Segment Detector (LSD)\footnote{\label{foot:lsd}\url{http://www.ipol.im/pub/art/2012/gjmr-lsd/}} \cite{VonGioi2010}. The source codes of  compared methods are obtained from the authors provide links.
	It is noticeable that the authors of Deep Wireframe Parser do not provide the pre-trained model for line segment detection, we reproduced their result by ourselves. 
	
	\begin{figure}[t!]
		\centering
		\includegraphics[width=0.7\linewidth]{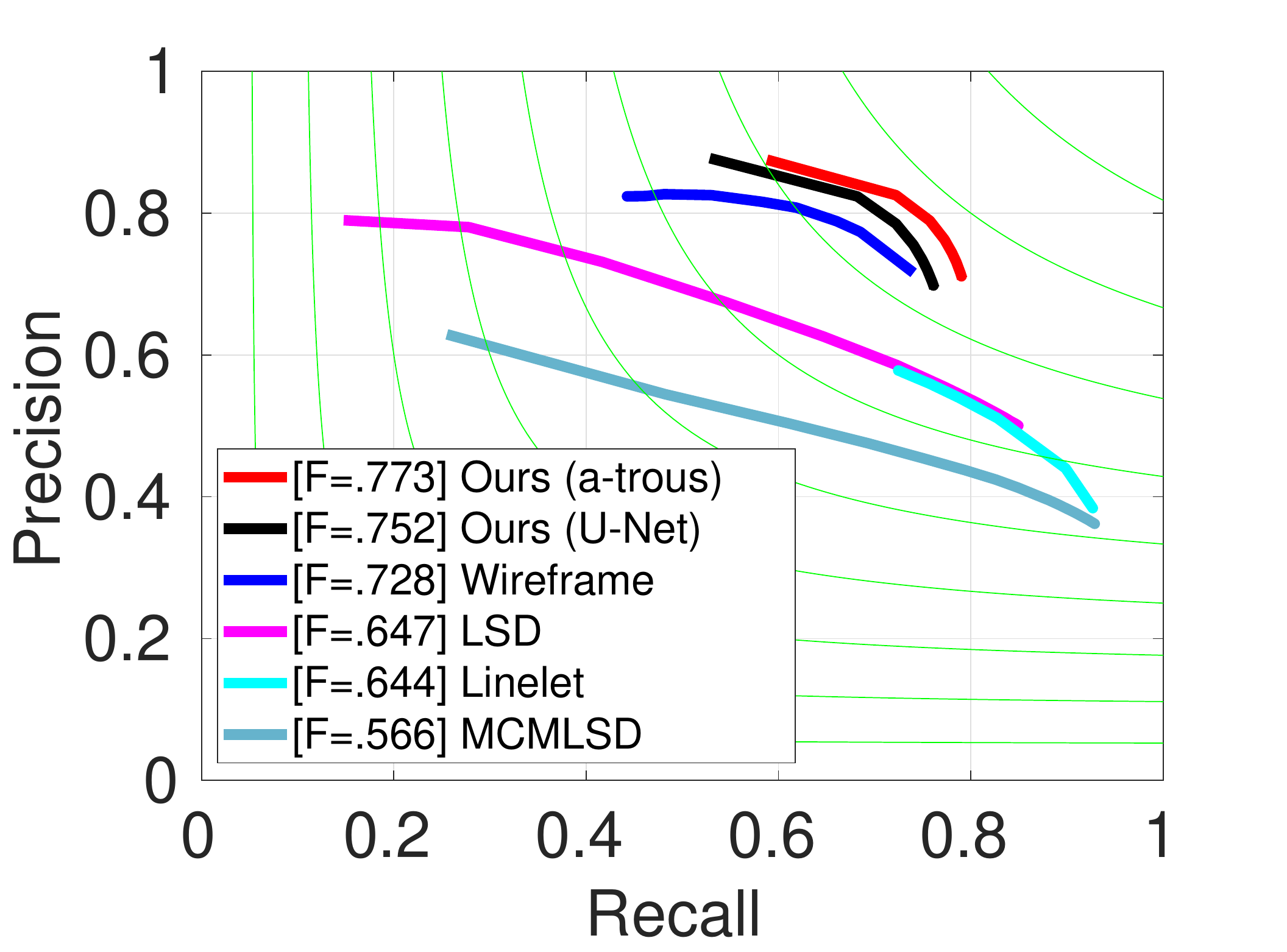}
		\caption{The PR curves of different line segment detection methods on the WireFrame~\cite{Huang2018a} dataset.}
		\label{fig:pr-curves-wireframe}
	\end{figure}
	\begin{figure}[t!]
		\centering
		\includegraphics[width=0.7\linewidth]{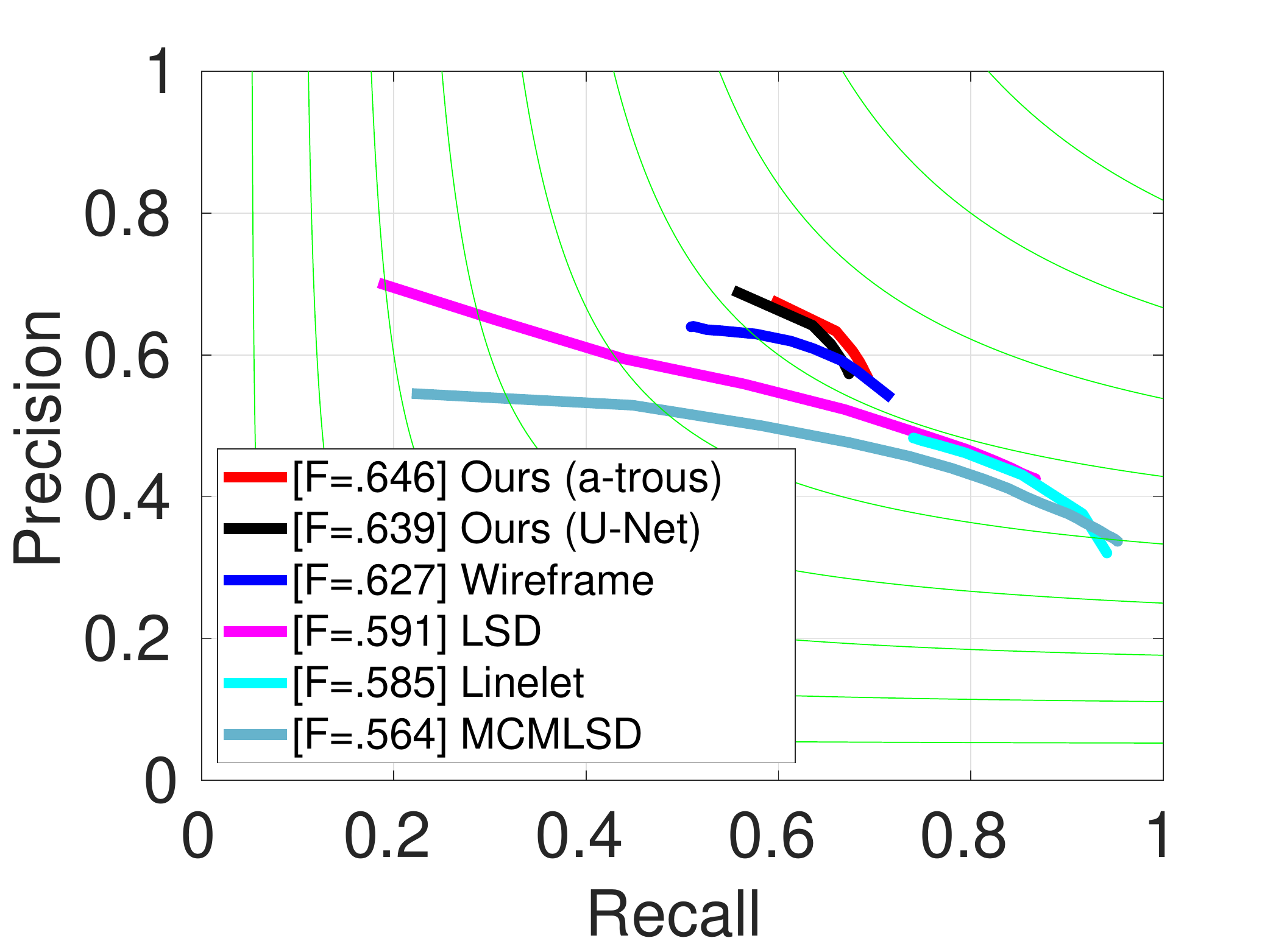}
		\caption{The PR curves of different line segment detection methods on the YorkUrban  datasets~\cite{Denis2008}.}
		\label{fig:pr-curves-york}
	\end{figure}
	\paragraph{Threshold Configuration}
	In our proposed method, we finally use the \emph{aspect ratio} to filter out false detections. Here, we vary the threshold of the \emph{aspect ratio} in the range $(0,1]$ with the step size $\Delta \tau = 0.1$. For comparison, the LSD is implemented with the $-\log(\text{NFA})$ in $0.01\times\{1.75^0, \ldots, 1.75^{19}\}$ where $\text{NFA}$ is the number of false alarm. Besides, Linelet \cite{Cho2018} use the same thresholds as the LSD to filter out false detection. For the MCMLSD \cite{Almazan_2017_CVPR}, we use the top $K$ detected line segments for comparison. 
	Due to the architecture of  Deep Wireframe Parser \cite{Huang2018a}, both the threshold for the junction localization confidence and the orientation confidence of junctions branches are fixed to $0.5$. Then, we use the author recommended threshold array $[2, 6, 10, 20, 30, 50, 80, 100, 150, 200, 250, 255]$ to binarize the line heat map and detect line segments.
	\begin{table}[t!]
		\centering
		\caption{F-measure evaluation with state-of-the-art approaches on the WireFrame dataset and York Urban dataset. The last column reports the averaged speed of different methods in \emph{frames per second} (FPS) on the WireFrame dataset.}
		\label{tab:fmeasure-and-fps}
		\begin{tabular}{l|c|c|c}
			\hline
			Methods & \makecell{Wireframe\\ dataset} & \makecell{York Urban\\ dataset}  & FPS\\
			\hline
			LSD \cite{VonGioi2010} & 0.647 & 0.591 & 19.6 \\
			MCMLSD \cite{Almazan_2017_CVPR} & 0.566 & 0.564 & 0.2 \\
			Linelet \cite{Cho2018} & 0.644 & 0.585 & 0.14\\
			Wireframe parser \cite{Huang2018a} & 0.728 & 0.627 & 2.24 \\
			\Xhline{1.pt}
			Ours (U-Net) & {\bf 0.752} & {\bf 0.639} & {\bf10.3} \\
			Ours (\emph{a-trous}) & {\bf 0.773} & {\bf 0.646} & {\bf 6.6} \\
			\hline
		\end{tabular}
	\end{table}
	\paragraph{Precision \& Recall}
	To compare our method with state-of-the-arts \cite{Huang2018a, Cho2018, Almazan_2017_CVPR, VonGioi2010},
	we evaluate the proposed method on the Wireframe dataset \cite{Huang2018a} and YorkUrban dataset \cite{Denis2008}. The precision-recall curves and F-measure are reported in Figure~\ref{fig:pr-curves-wireframe}, Figure~\ref{fig:pr-curves-york} and Table~ \ref{tab:fmeasure-and-fps}. Without bells and whistles, our proposed method outperforms all of these approaches on Wireframe and YorkUrban  datasets by a significant margin even with a 18-layer network. Deeper network architecture with ASPP module further improves the F-measure performance.
	Due to the YorkUrban dataset aiming at Manhattan frame estimation, some line segments in the images are not labeled, which causes the F-measure performance of all methods on this dataset decreased.
	
	\paragraph{Speed}
	We evaluate the computational time consuming for the abovementioned approaches on the Wireframe dataset. We run 462 frames with image reading and result writing steps and count the averaged time consuming because the size of testing images are not equal. As reported in Table \ref{tab:fmeasure-and-fps}, our proposed method can detect line segments fast (outperforms all methods except for the LSD) while getting the best performances. 
	All experiments perform on a PC workstation, which is equipped with an Intel Xeon E5-2620 $2.10$ GHz CPU and $4$ NVIDIA Titan X GPU devices. Only one GPU is used and the CPU programs are executed in a single thread. 
	
	Benefiting from the simplicity of original U-Net,  our method can detect line segments fast. The deep wireframe parser \cite{Huang2018a} spends much time for junction and line map fusion.
	On the other hand, benefiting from our novel formulation, we can resize the input images into $320\times 320$ and then transform the output line segments to the original scales, which can further reduce the computational cost.
	
	\begin{figure*}[t!]
		\centering
		\newcommand\outsize{0.14\linewidth}
		\def\wireframeA{00037538,00079421,00071820,00035326,00037075,00051663,00051831}
		\insertResults{\wireframeA}{lsd}{\outsize}{\small LSD}
		\insertResults{\wireframeA}{mcmlsd}{\outsize}{\small MCMLSD}
		\insertResults{\wireframeA}{linelet}{\outsize}{\small Linelet}
		\insertResults{\wireframeA}{wireframe}{\outsize}{\small Wireframe}
		\insertResults{\wireframeA}{ours}{\outsize}{\small Ours}
		\insertResults{\wireframeA}{gt}{\outsize}{\small GT}
		\caption{Some Results of line segment detection on  Wireframe \cite{Huang2018a} dataset with different approaches LSD \cite{VonGioi2010}, MCMLSD \cite{Almazan_2017_CVPR}, Linelet \cite{Cho2018}, Deep Wireframe Parser \cite{Huang2018a} and ours with the \emph{a-trous} Residual U-Net are shown from left to right. The ground truths are listed in last column as reference.}
		\vspace{-4mm}
		\label{fig:results_wireframe}
	\end{figure*}
	\begin{figure*}[t!]
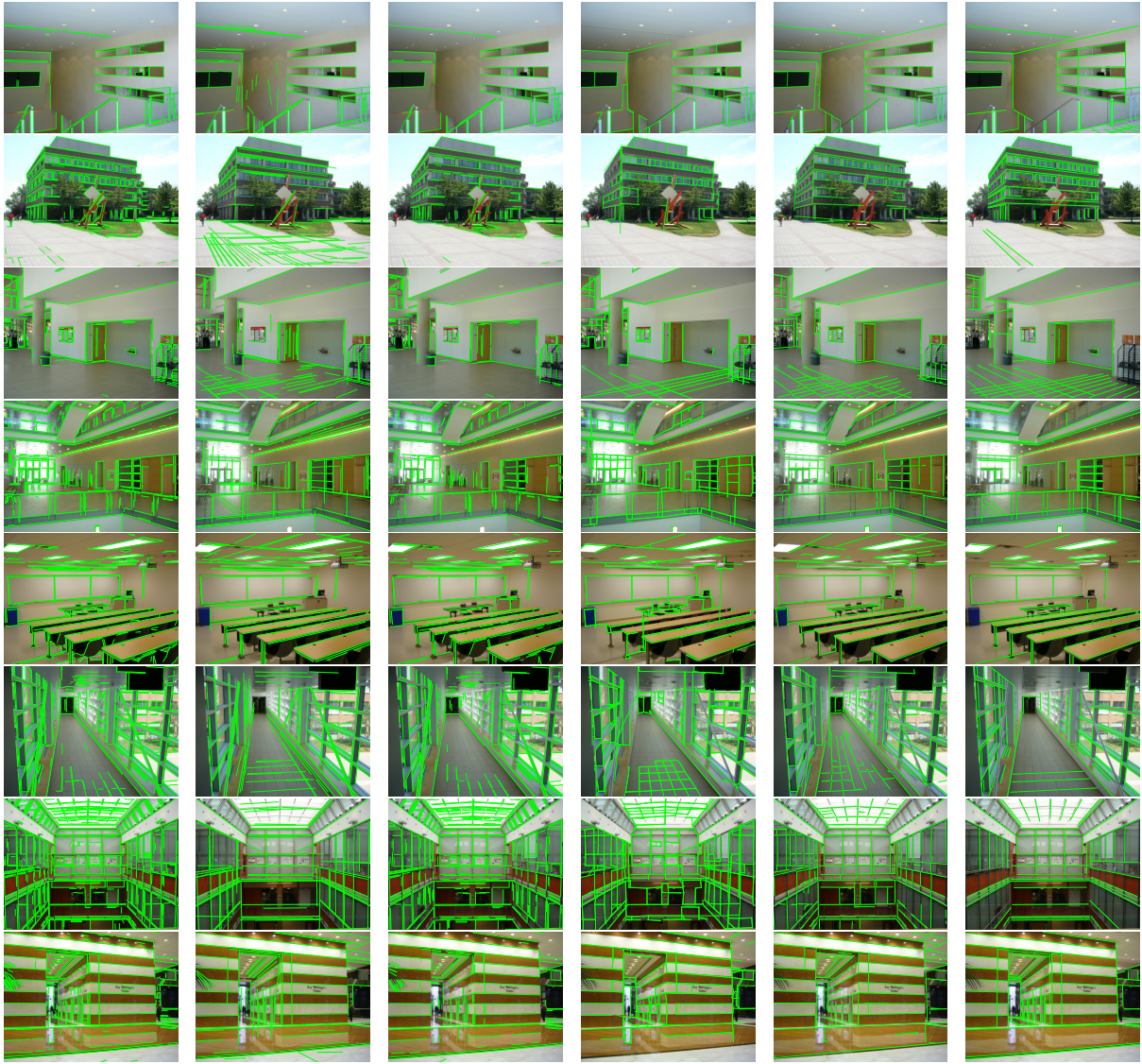

		\centering
		\newcommand\outsize{0.14\linewidth}
		\def\wireframeA{P1020825,P1020861,P1020824,P1020830,P1020838,P1020912,P1040833,P1080047}
		\insertResults{\wireframeA}{lsd}{\outsize}{\small LSD}
		\insertResults{\wireframeA}{mcmlsd}{\outsize}{\small MCMLSD}
		\insertResults{\wireframeA}{linelet}{\outsize}{\small Linelet}
		\insertResults{\wireframeA}{wireframe}{\outsize}{\small Wireframe}
		\insertResults{\wireframeA}{ours}{\outsize}{\small Ours}
		\insertResults{\wireframeA}{gt}{\outsize}{\small GT}
		\caption{Some Results of line segment detection on  YorkUrban \cite{Denis2008} dataset with different approaches LSD \cite{VonGioi2010}, MCMLSD \cite{Almazan_2017_CVPR}, Linelet \cite{Cho2018}, Deep Wireframe Parser \cite{Huang2018a} and ours with the \emph{a-trous} Residual U-Net are shown from left to right.  The ground truths are listed in last column as reference.}
		\label{fig:results_york}
	\end{figure*}
	\begin{figure*}[t!]
		\centering
		\newcommand\outsize{0.14\linewidth}
		\def\wireframeA{0004,0011,0015,0025,0027,0032,0048}
		\insertResults{\wireframeA}{lsd}{\outsize}{\small LSD}
		\insertResults{\wireframeA}{mcmlsd}{\outsize}{\small MCMLSD}
		\insertResults{\wireframeA}{linelet}{\outsize}{\small Linelet}
		\insertResults{\wireframeA}{wireframe}{\outsize}{\small Wireframe}
		\insertResults{\wireframeA}{ours}{\outsize}{\small Ours}
		\insertResults{\wireframeA}{rgb}{\outsize}{\small Input}
		\caption{Some Results of line segment detection on images fetched from Internet with different approaches LSD \cite{VonGioi2010}, MCMLSD \cite{Almazan_2017_CVPR}, Linelet \cite{Cho2018}, Deep Wireframe Parser \cite{Huang2018a} and ours with the \emph{a-trous} Residual U-Net are shown from left to right. The input images are listed in last column as reference.}
		\label{fig:results_internet}
	\end{figure*}
	
	\paragraph{Visualization}
	Further, we visualize the detected line segments with different methods on Wireframe dataset (see Figure~\ref{fig:results_wireframe}), YorkUrban dataset (see Figure~\ref{fig:results_york}) and images fetched from Internet (see Figure~\ref{fig:results_internet}). 
	Since the images fetched from Internet do not have ground truth annotations, we display the input images as reference for comparasion.
	The threshold configurations for visualization are as follow:
	\begin{enumerate}
		\item The \emph{a-contrario} validation of LSD and Linelet are set as $-\log\epsilon = 0.01\cdot1.75^{8}$;
		\item The top $90$ detected line segments for the MCMLSD are visualized;
		\item The threshold for line heat map is $10$ for the deep wireframe parser;
		\item The upper bound of aspect ratio is set as $0.2$ for our results.
	\end{enumerate}
	
	By observing these figures, it is easy to find that
	Deep Wireframe Parser \cite{Huang2018a} can detect more complete line segments compared with the previous methods, however, our proposed approach can get better result in the perspective of completeness. On the other hand, this junction driven approach indeed induces some uncertainty for the detection. The orientation of line segments estimated by junction branches is not accurate, which will affect the orientation of line segments. Meanwhile, some junctions are misconnected to get false detections.
	In contrast, our proposed method gets rid of junction detection and directly detect the line segments from images. 
	
	Comparing with the rest of approaches \cite{VonGioi2010,Almazan_2017_CVPR,Cho2018},
	the deep learning based methods (including ours) can utilize the global information to get complete results in the low-contrast regions while suppressing the false detections in the edge-like texture regions. Due to the limitation of local features, the approaches \cite{VonGioi2010,Almazan_2017_CVPR,Cho2018} cannot handle the results with global information and still get some false detections even with powerful validation approaches. 
	Although the overall F-measure of LSD is slightly better than Linelet, the visualization results of Linelet are cleaner.
	
	\section{Conclusion}
	\label{sec:conclusion}
	In this paper, we proposed a method of building the duality between the region-partition based attraction field representation and the line segment representation. We then pose the problem of line segment detection (LSD) as the region coloring problem which is addressed by learning convolutional neural networks. The proposed attraction field representation rigorously addresses several challenges in LSD such as local ambiguity and class imbalance. The region coloring formulation of LSD harnesses the best practices developed in ConvNets based semantic segmentation methods such as the encoder-decoder architecture and the {\em a-trous} convolution. In experiments, our method is tested on two widely used LSD benchmarks, the WireFrame dataset~\cite{Huang2018a} and the YorkUrban dataset~\cite{Denis2008}, with state-of-the-art performance obtained and $6.6\sim 10.4$ FPS speed. 
	
	{
		\bibliographystyle{ieeetr}
		\bibliography{ref}
	}
\end{document}

%% file: figures/support_region.tex
\begin{tikzpicture}[scale=0.2]
        \pgfmathsetmacro{\ymin}{0}
        \pgfmathsetmacro{\xmin}{0}
        \pgfmathsetmacro{\ymax}{10}
        \pgfmathsetmacro{\xmax}{10}	
        \node[circle, fill, inner sep=1.0pt, red] (A0) at (2.36,8.1) {};
        \node[circle, fill, inner sep=1.0pt, red] (A1) at (7.23,6.83) {};
        \node[circle, fill, inner sep=1.0pt, green] (B0) at (8,1.2) {};
        \node[circle, fill, inner sep=1.0pt, green] (B1) at (8.7,6.5) {};
        \node[circle, fill, inner sep=1.0pt, blue] (C0) at (1.73,4.29) {};
        \node[circle, fill, inner sep=1.0pt, blue] (C1) at (5.60,1.40) {};
        
        \draw [blue, very thick] (A0) -- (A1);	
        \draw [green, very thick] (B0) -- (B1);
        \draw [red, very thick] (C0) -- (C1);

        \foreach \i in {\xmin,\xmax} {
			\draw [thick, gray] (\i, \ymin) -- (\i, \ymax) node [below] at (\i, \ymin) {};
		}
		\foreach \i in {\ymin,\ymax} {
		\draw [thick, gray] (\xmin, \i) -- (\xmax, \i) node [below] at (\xmin, \i) {};
	    }        
        \draw [step=1.0, black] (\xmin, \ymin) grid (\xmax,\ymax);

        \fill [green, opacity=0.2] (0,0) -- (7,0) -- (7,3) -- (6,3) -- (6,5) -- (4,5) -- (4,6) -- (0,6) -- (0,0);
        \fill [red, opacity=0.2] (6,5) -- (6,3) -- (7,3) -- (7,0) -- (10,0) -- (10,10) -- (9,10) -- (9,9) -- (8,9) -- (8,8) -- (8,7) -- (8,6) -- (7,6) -- (7,5) -- (6,5);
        \fill [blue, opacity=0.2] (0,6) -- (4,6) -- (4,5) -- (7,5) -- (7,6) -- (8,6) -- (8,9) -- (9,9) -- (9,10) -- (0,10) -- (0,6);
        \end{tikzpicture}

%% file: figures/offset_details.tex
\begin{tikzpicture}[scale=0.33]
\pgfmathsetmacro{\ymin}{0}
\pgfmathsetmacro{\xmin}{0}
\pgfmathsetmacro{\ymax}{6}
\pgfmathsetmacro{\xmax}{7}	
\foreach \i in {\xmin,...,\xmax} {
		\draw [thick, gray] (\i, \ymin) -- (\i, \ymax) node [below] at (\i, \ymin) {};
	}
\foreach \i in {\ymin,\ymax} {
	\draw [thick, gray] (\xmin, \i) -- (\xmax, \i) node [below] at (\xmin, \i) {};
    }
\draw [step=1.0, black] (\xmin, \ymin) grid (\xmax,\ymax);
\node[circle, fill, inner sep=1.0pt, blue] (A) at (1.73,4.29) {};
\node[circle, fill, inner sep=1.0pt, blue] (B) at (5.60,1.40) {};
\draw [red, very thick] (A) -- (B);
\fill [green, opacity=0.2] (0,0) -- (7,0) -- (7,3) -- (6,3) -- (6,5) -- (4,5) -- (4,6) -- (0,6) -- (0,0);

\node[circle, fill, inner sep=0.5pt, gray](ea1) at (0.5,5.5) {};
\node[circle, fill, inner sep=0.5pt, gray](ea2) at (1.5,5.5) {};
\node[circle, fill, inner sep=0.5pt, gray](ea3) at (0.5,4.5) {};
\node[circle, fill, inner sep=0.5pt, gray](ea4) at (1.5,4.5) {};
\node[circle, fill, inner sep=0.5pt, gray](ea5) at (0.5,3.5) {};
\node[circle, fill, inner sep=0.5pt, blue](ea6) at (2.5,5.5) {};
\node[circle, fill, inner sep=0.5pt, gray](eb1) at (6.5,2.5) {};
\node[circle, fill, inner sep=0.5pt, gray](eb2) at (6.5,1.5) {};
\node[circle, fill, inner sep=0.5pt, gray](eb3) at (6.5,0.5) {};
\node[circle, fill, inner sep=0.5pt, gray](eb4) at (5.5,0.5) {};
\node[circle, fill, inner sep=0.5pt, blue](c3) at (1.5,1.5) {};
\node[circle, fill, inner sep=0.5pt, blue](d2) at (2.5,1.5) {};
\node[circle, fill, inner sep=0.5pt, blue](e4) at (3.5,3.5) {};
\node[circle, fill, inner sep=0.5pt, blue](f1) at (4.5,0.5) {};
\node[circle, fill, inner sep=0.5pt, blue](f2) at (4.5,1.5) {};	
\node[circle, fill, inner sep=0.5pt, blue](g3) at (5.5,4.5)  {};	
\draw [->,>=stealth, dashed, thick] (ea1) -- (A);
\draw [->,>=stealth, dashed, thick] (ea2) -- (A);
\draw [->,>=stealth, dashed, thick] (ea3) -- (A);
\draw [->,>=stealth, dashed, thick] (ea4) -- (A);
\draw [->,>=stealth, dashed, thick] (ea5) -- (A);
\draw [->,>=stealth, dashed, thick] (ea6) -- (A);
\draw [->,>=stealth, dashed, thick] (eb1) -- (B);
\draw [->,>=stealth, dashed, thick] (eb2) -- (B);
\draw [->,>=stealth, dashed, thick] (eb3) -- (B);
\draw [->,>=stealth, dashed, thick] (eb4) -- (B);
\draw [->,>=stealth, thick] (c3) -- ($(A)!(c3)!(B)$);
\draw [->,>=stealth, thick] (d2) -- ($(A)!(d2)!(B)$);
\draw [->,>=stealth, thick] (e4) -- ($(A)!(e4)!(B)$);
\draw [->,>=stealth, thick] (f1) -- ($(A)!(f1)!(B)$);
\draw [->,>=stealth, thick] (f2) -- ($(A)!(f2)!(B)$);
\draw [->,>=stealth, thick] (g3) -- ($(A)!(g3)!(B)$);
\end{tikzpicture}

%% file: figures/squeeze.tex
\begin{tikzpicture}[scale=0.33]
\pgfmathsetmacro{\ymin}{0}
\pgfmathsetmacro{\xmin}{0}
\pgfmathsetmacro{\ymax}{6}
\pgfmathsetmacro{\xmax}{7}	
\foreach \i in {\xmin,...,\xmax} {
		\draw [thick, gray, opacity=0] (\i, \ymin) -- (\i, \ymax) node [below] at (\i, \ymin) {};
	}
\foreach \i in {\ymin,\ymax} {
	\draw [thick, gray] (\xmin, \i) -- (\xmax, \i) node [below] at (\xmin, \i) {};
    }
\draw [step=1.0, black] (\xmin, \ymin) grid (\xmax,\ymax);
\node[inner sep=1.0pt, blue] (A) at (1.73,4.29) {};
\node[inner sep=1.0pt, blue] (B) at (5.60,1.40) {};
\coordinate (dir) at (-0.26708016,  0.19944745);
\node[inner sep=0.5pt] (a0) at (0.5, 0.5) {};
\node[inner sep=0.5pt] (a0p) at ($(A)!(a0)!(B)$){};
\draw[->] (a0p) -- ++(dir);
\node[inner sep=0.5pt] (a1) at (1.5, 0.5){};
\node[inner sep=0.5pt] (a1p) at ($(A)!(a1)!(B)$){};
\draw[->] (a1p) -- ++(dir);
\node[inner sep=0.5pt] (a2) at (2.5, 0.5){};
\node[inner sep=0.5pt] (a2p) at ($(A)!(a2)!(B)$){};
\draw[->] (a2p) -- ++(dir);
\node[inner sep=0.5pt] (a3) at (3.5, 0.5){};
\node[inner sep=0.5pt] (a3p) at ($(A)!(a3)!(B)$){};
\draw[->] (a3p) -- ++(dir);
\node[inner sep=0.5pt] (a4) at (4.5, 0.5){};
\node[inner sep=0.5pt] (a4p) at ($(A)!(a4)!(B)$){};
\draw[->] (a4p) -- ++(dir);
\node[inner sep=0.5pt] (a5) at (0.5, 1.5){};
\node[inner sep=0.5pt] (a5p) at ($(A)!(a5)!(B)$){};
\draw[->] (a5p) -- ++(dir);
\node[inner sep=0.5pt] (a6) at (1.5, 1.5){};
\node[inner sep=0.5pt] (a6p) at ($(A)!(a6)!(B)$){};
\draw[->] (a6p) -- ++(dir);
\node[inner sep=0.5pt] (a7) at (2.5, 1.5){};
\node[inner sep=0.5pt] (a7p) at ($(A)!(a7)!(B)$){};
\draw[->] (a7p) -- ++(dir);
\node[inner sep=0.5pt] (a8) at (3.5, 1.5){};
\node[inner sep=0.5pt] (a8p) at ($(A)!(a8)!(B)$){};
\draw[->] (a8p) -- ++(dir);
\node[inner sep=0.5pt] (a9) at (4.5, 1.5){};
\node[inner sep=0.5pt] (a9p) at ($(A)!(a9)!(B)$){};
\draw[->] (a9p) -- ++(dir);
\node[inner sep=0.5pt] (a10) at (0.5, 2.5){};
\node[inner sep=0.5pt] (a10p) at ($(A)!(a10)!(B)$){};
\draw[->] (a10p) -- ++(dir);
\node[inner sep=0.5pt] (a11) at (1.5, 2.5){};
\node[inner sep=0.5pt] (a11p) at ($(A)!(a11)!(B)$){};
\draw[->] (a11p) -- ++(dir);
\node[inner sep=0.5pt] (a12) at (2.5, 2.5){};
\node[inner sep=0.5pt] (a12p) at ($(A)!(a12)!(B)$){};
\draw[->] (a12p) -- ++(dir);
\node[inner sep=0.5pt] (a13) at (3.5, 2.5){};
\node[inner sep=0.5pt] (a13p) at ($(A)!(a13)!(B)$){};
\draw[->] (a13p) -- ++(dir);
\node[inner sep=0.5pt] (a14) at (4.5, 2.5){};
\node[inner sep=0.5pt] (a14p) at ($(A)!(a14)!(B)$){};
\draw[->] (a14p) -- ++(dir);
\node[inner sep=0.5pt] (a15) at (5.5, 2.5){};
\node[inner sep=0.5pt] (a15p) at ($(A)!(a15)!(B)$){};
\draw[->] (a15p) -- ++(dir);
\node[inner sep=0.5pt] (a16) at (1.5, 3.5){};
\node[inner sep=0.5pt] (a16p) at ($(A)!(a16)!(B)$){};
\draw[->] (a16p) -- ++(dir);
\node[inner sep=0.5pt] (a17) at (2.5, 3.5){};
\node[inner sep=0.5pt] (a17p) at ($(A)!(a17)!(B)$){};
\draw[->] (a17p) -- ++(dir);
\node[inner sep=0.5pt] (a18) at (3.5, 3.5){};
\node[inner sep=0.5pt] (a18p) at ($(A)!(a18)!(B)$){};
\draw[->] (a18p) -- ++(dir);
\node[inner sep=0.5pt] (a19) at (4.5, 3.5){};
\node[inner sep=0.5pt] (a19p) at ($(A)!(a19)!(B)$){};
\draw[->] (a19p) -- ++(dir);
\node[inner sep=0.5pt] (a20) at (5.5, 3.5){};
\node[inner sep=0.5pt] (a20p) at ($(A)!(a20)!(B)$){};
\draw[->] (a20p) -- ++(dir);
\node[inner sep=0.5pt] (a21) at (1.5, 4.5){};
\node[inner sep=0.5pt] (a21p) at ($(A)!(a21)!(B)$){};
\draw[->] (a21p) -- ++(dir);
\node[inner sep=0.5pt] (a22) at (2.5, 4.5){};
\node[inner sep=0.5pt] (a22p) at ($(A)!(a22)!(B)$){};
\draw[->] (a22p) -- ++(dir);
\node[inner sep=0.5pt] (a23) at (3.5, 4.5){};
\node[inner sep=0.5pt] (a23p) at ($(A)!(a23)!(B)$){};
\draw[->] (a23p) -- ++(dir);
\node[inner sep=0.5pt] (a24) at (4.5, 4.5){};
\node[inner sep=0.5pt] (a24p) at ($(A)!(a24)!(B)$){};
\draw[->] (a24p) -- ++(dir);
\node[inner sep=0.5pt] (a25) at (5.5, 4.5){};
\node[inner sep=0.5pt] (a25p) at ($(A)!(a25)!(B)$){};
\draw[->] (a25p) -- ++(dir);
\draw [red,opacity=0.5,->] (A) -- ++(-0.35643939549437625, 0.3506436329660124);
\draw [red,opacity=0.5,->] (A) -- ++(-0.09336950361691489, 0.4912047798976827);
\draw [red,opacity=0.5,->] (A) -- ++(-0.4928682127552036, 0.08414823144601036);
\draw [red,opacity=0.5,->] (A) -- ++(-0.36924274696406295, 0.33713468201066615);
\draw [red,opacity=0.5,->] (A) -- ++(-0.42070033194990575, -0.2702059042605086);
\draw [red,opacity=0.5,->] (A) -- ++(0.26843774609657967, 0.42183074386605374);
\draw [red,opacity=0.5,->] (5.6, 1.4) -- ++(-0.3166188951286314, -0.38697864960166045);
\draw [red,opacity=0.5,->] (5.6, 1.4) -- ++(-0.4969418673368095, -0.0552157630374233);
\draw [red,opacity=0.5,->] (5.6, 1.4) -- ++(-0.35355339059327384, 0.3535533905932737);
\draw [red,opacity=0.5,->] (5.6, 1.4) -- ++(0.055215763037423087, 0.4969418673368095);
\fill[green,opacity=0.2] (1.0, 4.0) rectangle (2.0, 5.0);
\fill[green,opacity=0.2] (2.0, 4.0) rectangle (3.0, 5.0);
\fill[green,opacity=0.2] (2.0, 3.0) rectangle (3.0, 4.0);
\fill[green,opacity=0.2] (3.0, 3.0) rectangle (4.0, 4.0);
\fill[green,opacity=0.2] (3.0, 2.0) rectangle (4.0, 3.0);
\fill[green,opacity=0.2] (4.0, 2.0) rectangle (5.0, 3.0);
\fill[green,opacity=0.2] (4.0, 1.0) rectangle (5.0, 2.0);
\fill[green,opacity=0.2] (5.0, 1.0) rectangle (6.0, 2.0);
\end{tikzpicture}